\documentclass[10pt, conference, compsocconf]{IEEEtran}

\usepackage{graphicx}
\usepackage{amssymb}
\usepackage{amsthm}
\usepackage{amsmath}
\usepackage{amsfonts}
\usepackage{subfigure}
\usepackage{lineno}
\usepackage{multirow}
\usepackage{threeparttable}
\usepackage{mathrsfs}
\usepackage{amsbsy}
\usepackage{url}

\usepackage[noend]{algpseudocode}
\usepackage{algorithmicx}
\usepackage{algorithm}

\newtheorem{definition}{Definition}

\newtheorem{example}{Example}
\ifCLASSINFOpdf

\else

\fi

\begin{document}
\title{Fast Modeling Methods for Complex System with Separable Features}

\author{\IEEEauthorblockN{Chen Chen\IEEEauthorrefmark{2}\IEEEauthorrefmark{3},
Changtong Luo\IEEEauthorrefmark{1}\IEEEauthorrefmark{2},
Zonglin Jiang\IEEEauthorrefmark{2}\IEEEauthorrefmark{3}}
\IEEEauthorblockA{\IEEEauthorrefmark{2}State Key Laboratory of High Temperature Gas Dynamics, Institute of Mechanics \\ Chinese Academy of Sciences, Beijing 100190, China}
\IEEEauthorblockA{\IEEEauthorrefmark{3}School of Engineering Sciences, University of Chinese Academy of Sciences, Beijing\\ 100049, China}
\IEEEauthorrefmark{1}Email: luo@imech.ac.cn}

\maketitle
\begin{abstract}
Data-driven modeling plays an increasingly important role in different areas of engineering. For most of existing methods, such as genetic programming (GP), the convergence speed might be too slow for large scale problems with a large number of variables. Fortunately, in many applications, the target models are separable in some sense. In this paper, we analyze different types of separability of some real-world engineering equations and establish a mathematical model of generalized separable system (GS system). In order to get the structure of the GS system, two concepts, namely `block' and `factor' are introduced, and a special method, block and factor detection is also proposed, in which the target model is decomposed into a number of blocks, further into minimal blocks and factors. Compare to the conventional GP, the new method can make large reductions to the search space. The minimal blocks and factors are optimized and assembled with a global optimization search engine, low dimensional simplex evolution (LDSE). An extensive study between the proposed method and a state-of-the-art data-driven fitting tool, Eureqa, has been presented with several man-made problems. Test results indicate that the proposed method is more effective and efficient under all the investigated cases.

\end{abstract}

\IEEEpeerreviewmaketitle
\section{Introduction}
Data-driven modeling has become a powerful technique in different areas of engineering, such as industrial data analysis \cite{Luo2015}, circuits analysis and design \cite{Shokouhifar2015}, signal processing \cite{Yang2005}, system identification \cite{Guo2012}, etc. For a concerned data-driven modeling problem, we aim to find a performance function that best explains the relationship between input variables and the target system (or constrained system) based on a given set of sample points. Among the existing methods, genetic programming (GP) \cite{Koza1992} is a classical approach. Theoretically, GP can get an optimal solution provided that the computation time is long enough. However, the computational cost of GP for a large scale problem with a large number of input variables is still very expensive.

In many scientific or engineering problems, the target model are separable. Luo et al. \cite{Luo2017} have presented a divide-and-conquer (D\&C) method for GP. The authors indicated that the solving process could be accelerated by detecting the correlation between each variable and the target function. In \cite{Luo2017}, a special method, bi-correlation test (BiCT), was proposed to divide a concerned target function into a number of sub-functions. Compared to conventional GP, D\&C method could reduce the computational effort (computational complexity) by orders of magnitude.

In this paper, different types of separability of some practical engineering problems are analyzed, and a mathematical model of generalized separable system (GS system) is established. In order to get the structure of the GS system, a block and factor detection method is proposed, where the target model is decomposed into a number of block, further into minimal blocks and factors. The new method is an improved version of the BiCT method \cite{Luo2017}. The performance of the proposed method is compared with the results of Eureqa, which is a state-of-the-art data-driven fitting tool. Numerical results show that the proposed method is effective, and is able to recover all the investigated cases rapidly and reliably.

\section{Observation and discussion of the types of separability}
\label{Section2}

\subsection{Observation}
\label{Section2-1}
In this section, three examples of real-world problems are given as follows to illustrate several common types of separability in practical problems.

\begin{example}
\label{ex2-1}
When developing a rocket engine, it is crucial to model the internal flow of a high-speed compressible gas through the nozzle. The closed-form expression for the mass flow through a choked nozzle is
\begin{equation}
\label{nozzle}
\dot m = \frac{{{p_0}{A^*}}}{{\sqrt {{T_0}} }}\sqrt {\frac{\gamma }{R}{{\left( {\frac{2}{{\gamma  + 1}}} \right)}^{{{\left( {\gamma  + 1} \right)} \mathord{\left/
 {\vphantom {{\left( {\gamma  + 1} \right)} {\left( {\gamma  - 1} \right)}}} \right.
 \kern-\nulldelimiterspace} {\left( {\gamma  - 1} \right)}}}}}.
\end{equation}
In Eq. (\ref{nozzle}), the five independent variables, $p_0$, $T_0$, $A^*$, $R$ and $\gamma$ are all separable. The equation can be called a multiplicatively separable function, which can be re-expressed as follows
\begin{equation}
\label{nozzle2}
\begin{aligned}
\dot m & = f\left( {{p_0},{A^*},{T_0},R,\gamma } \right) \\
& = {\varphi _1}\left( {{p_0}} \right) \times {\varphi _2}\left( {{A^*}} \right) \times {\varphi _3}\left( {{T_0}} \right) \times {\varphi _4}\left( R \right) \times {\varphi _5}\left( \gamma  \right).
\end{aligned}
\end{equation}
\end{example}

\begin{example}
\label{ex2-2}
In aircraft design, the lift coefficient of a whole aircraft can be expressed as
\begin{equation}
\label{CL}
{C_L} = {C_{L\alpha }}\left( {\alpha  - {\alpha_0}} \right) + {C_{L{\delta _e}}}{\delta _e}\frac{{{S_{{\text{HT}}}}}}{{{S_{{\text{ref}}}}}},
\end{equation}
where the variable $C_{L\alpha }$, $C_{L{\delta _e}}$, $\delta _e$, $S_{{\text{HT}}}$ and $S_{{\text{ref}}}$ are separable. The variable $\alpha$ and ${\alpha_0}$ are not separable, but their combination $\left( \alpha,\alpha_0 \right)$ can be considered separable. Hence, Eq. (\ref{CL}) can be re-expressed as
\begin{equation}
\label{CL2}
\begin{aligned}
{C_L}  =&  f\left( {{C_{L\alpha }},\alpha ,{\alpha _0},{C_{L{\delta _e}}},{\delta _e},{S_{{\text{HT}}}},{S_{{\text{ref}}}}} \right) \\
 =& {\varphi _1}\left( {{C_{L\alpha }}} \right) \times {\varphi _2}\left( {\alpha ,{\alpha _0}} \right) \\ &+ {\varphi _3}\left( {{C_{L{\delta _e}}}} \right) \times {\varphi _4}\left( {{\delta _e}} \right)\times{\varphi _5}\left( {{S_{{\text{HT}}}}} \right) \times {\varphi _6}\left( {{S_{{\text{ref}}}}} \right).
\end{aligned}
\end{equation}
\end{example}

\begin{example}
\label{ex2-3}
The flow past a circular cylinder is a classical problem in fluid dynamics. A valid stream function for the inviscid, incompressible flow over a circular cylinder of radius $R$ is
\begin{equation}
\label{flowequ}
\psi = \left( {{V_\infty }r\sin \theta } \right)\left( {1 - \frac{{{R^2}}}{{{r^2}}}} \right) + \frac{\Gamma }{{2\pi }}\ln \frac{r}{R},
\end{equation}
which can be re-expressed as
\begin{equation}
\label{flowequ2}
\begin{aligned}
  \psi  =&   f\left( {{V_\infty },\sin \theta ,R,r,\Gamma } \right) \\ 
   =&   {\varphi _1}\left( {{V_\infty }} \right) \times {\varphi _2}\left( {\sin \theta } \right) \times \boxed{{\varphi _3}\left( {r,R} \right)} \\ &+ {\varphi _4}\left( \Gamma  \right) \times \boxed{{\varphi _5}\left( {r,R} \right)}. 
\end{aligned}
\end{equation}
Note that the variable $r$ and $R$ appear twice in Eq. (\ref{flowequ}). In other words, variable $r$ and $R$ have two sub-functions, namely ${\varphi _3}\left( {r,R} \right) = \left( {1 - {{{R^2}} \mathord{\left/
 {\vphantom {{{R^2}} {{r^2}}}} \right.
 \kern-\nulldelimiterspace} {{r^2}}}} \right) \cdot r$ and ${\varphi _5}\left( {r,R} \right) = \ln \left( {{r \mathord{\left/
 {\vphantom {r R}} \right.
 \kern-\nulldelimiterspace} R}} \right)$. Although Eq. (\ref{flowequ2}) is not a strictly separable function, the variables of Eq. (\ref{flowequ2}) also have separability.
\end{example}

\subsection{Discussion}
\label{Section2-2}
As seen from the above subsection, many practical problems have the feature of separability. Luo et al. suggested using the separability to accelerate the conventional GP for data-driven modeling. The separable function introduced in \cite{Luo2017} could be described as follows.
\begin{definition}[Separable function]
\label{def1}
A scalar function $f\left( X \right)$ with $n$ continuous variables $X = \left\{ {{x_i}:i = 1,2, \cdots ,n} \right\}$ ($f:{\mathbb{R}^n} \mapsto \mathbb{R}$, $X \subset \Omega \in {\mathbb{R}^n}$, where $\Omega$ is a closed bounded convex set, such that $\Omega  = \left[ {{a_1},{b_1}} \right] \times \left[ {{a_2},{b_2}} \right] \times  \cdots  \times \left[ {{a_n},{b_n}} \right]$) is said to be separable if and only if it can be written as
\begin{equation}
\label{def1eq}
f\left( X \right) = {c_0}{ \otimes _1}{c_1}{\varphi _1}\left( {{X_1}} \right){ \otimes _2}{c_2}{\varphi _2}\left( {{X_2}} \right){ \otimes _3} \cdots { \otimes _m}{c_m}{\varphi _m}\left( {{X_m}} \right),
\end{equation}
where the variable set ${X_i}$ is a proper subset of $X$, such that ${X_i} \subset X$ with $\bigcup\nolimits_{i = 1}^m {{X_i}}  = X$, $\bigcap\nolimits_{i = 1}^m {{X_i}}  = \emptyset$, and the cardinal number of ${X_i}$ is denoted by ${\text{card}}\left( {{X_i}} \right) = {n_i}$, for $\sum\nolimits_{i = 1}^m {{n_i}}  = n$ and $i = 1,2, \cdots ,m$. 
Sub-function ${\varphi _i}$ is a scalar function such that ${\varphi _i}:{\mathbb{R}^{{n_i}}} \mapsto \mathbb{R}$. 
The binary operator $\otimes_i$ could be plus ($+$) and times ($\times$).
\end{definition}

Note that binary operator, minus ($-$) and division ($/$), are not included in $\otimes$ for simplicity. This does not affect much of its generality, since minus ($-$) could be regarded as $\left(  -  \right) = \left( { - 1} \right) \cdot \left(  +  \right)$, and sub-function could be treated as ${\tilde \varphi _i}\left(  \cdot  \right) = {1 \mathord{\left/
 {\vphantom {1 {{\varphi _i}\left(  \cdot  \right)}}} \right.
 \kern-\nulldelimiterspace} {{\varphi _i}\left(  \cdot  \right)}}$ if only ${\varphi _i}\left(  \cdot  \right) \ne 0$.

We can see that Example \ref{ex2-3} is inconsistent with the above definition of the separable function. It is because that some variables (e.g., variable ${V_\infty }$, $\sin \theta$ and $\Gamma$ of Eq. (\ref{flowequ})) appears only once in a concerned target model, while the other variables (e.g., variable $r$ and $R$ of Eq. (\ref{flowequ})) appears more than once. This feature motivates us to generalize the mathematical form of the separable function, namely Eq. (\ref{def1eq}), and establish a more general model.

\section{The mathematical model of generalized separable system}
\label{Section3}

\begin{definition}[Generalized separable system]
\label{def-GSS}
The mathematical model of a generalized separable system $f\left( X \right)$ with $n$ continuous variables $X = \left\{ {{x_i}:i = 1,2, \cdots ,n} \right\}$, ($f:{\mathbb{R}^n} \mapsto \mathbb{R}$, $X \subset \Omega \in {\mathbb{R}^n}$, where $\Omega$ is a closed bounded convex set, such that $\Omega  = \left[ {{a_1},{b_1}} \right] \times \left[ {{a_2},{b_2}} \right] \times  \cdots  \times \left[ {{a_n},{b_n}} \right]$) is defined as
\begin{equation}
\label{eq-GSS}
\begin{aligned}
f\left( X \right) &= f\left( {{X^r},{{\bar X}^r}} \right) = {c_0} + \sum\limits_{i = 1}^m {{c_i}{\varphi _i}\left( {X_i^r,\bar X_i^r} \right)}  \\&= {c_0} + \sum\limits_{i = 1}^m {{c_i}{{\tilde \omega }_i}\left( {X_i^r} \right){{\tilde \psi }_i}\left( {\bar X_i^r} \right)}\\&= {c_0} + \sum\limits_{i = 1}^m {{c_i}\prod\limits_{j = 1}^{{p_i}} {{\omega _{i,j}}\left( {X_{i,j}^r} \right)} \prod\limits_{k = 1}^{{q_i}} {{\psi _{i,k}}\left( {\bar X_{i,k}^r} \right)} },
\end{aligned}
\end{equation}
where the variable set ${X^r} = \left\{ {{x_i}:i = 1,2, \cdots ,l} \right\}$ is a proper subset of $X$, such that ${X^r} \subset X$, and the cardinal number of ${X^r}$ is ${\text{card}}\left( {{{X}^r}} \right) = l$. 
${{\bar X}^r}$ is the complementary set of $X^r$ in $X$, i.e. ${{\bar X}^r} = {\complement _X}{X^r}$, where ${\text{card}}\left( {{{\bar X}^r}} \right) = n - l$. 
$X_i^r$ is the subset of ${X^r}$, such that $X_i^r \subseteq {X^r}$, where ${\text{card}}\left( {X_i^r} \right) = {r_i}$. 
$X_{i,j}^r \subseteq X_i^r$, 
such that $\bigcup\nolimits_{j = 1}^{{p_i}} {X_{i,j}^r}  = X_i^r$, $\bigcap\nolimits_{j = 1}^{{p_i}} {X_{i,j}^r}  = \emptyset$, where ${\text{card}}\left( {X_{i,j}^r} \right) = {r_{i,j}}$, for $i = 1,2, \cdots ,m$, $j = 1,2, \cdots ,{p_i}$ and $\sum\nolimits_{j = 1}^{{p_i}} {{r_{i,j}}}  = {r_i}$.
$\bar X_i^r \subset {{\bar X}^r}$ ($\bar X_i^r \ne \emptyset$), 
such that $\bigcup\nolimits_{i = 1}^m {\bar X_i^r}  = {{\bar X}^r}$, $\bigcap\nolimits_{i = 1}^m {\bar X_i^r}  = \emptyset$, where ${\text{card}}\left( {\bar X_i^r} \right) = {s_i}$, for $s_i \geqslant 1$, $\sum\nolimits_{i = 1}^m {{s_i}}  = n-l$.
$\bar X_{i,k}^r \subseteq \bar X_i^r$, 
such that $\bigcup\nolimits_{k = 1}^{{q_i}} {\bar X_{i,k}^r}  = \bar X_i^r$, $\bigcap\nolimits_{k = 1}^{{q_i}} {\bar X_i^r}  = \emptyset$, where ${\text{card}}\left( {\bar X_{i,k}^r} \right) = {s_{i,k}}$, for $k = 1,2, \cdots ,{q_i}$ and $\sum\nolimits_{k = 1}^{{q_i}} {{s_{i,k}}}  = {s_i}$. 
Sub-functions ${\varphi _i}$, ${{\tilde \omega }_i}$, ${{\tilde \psi }_i}$, ${\omega _{i,j}}$ and ${\psi _{i,k}}$ are scalar functions, 
such that ${\varphi _i}:{\mathbb{R}^{r_i+s_i}} \mapsto \mathbb{R}$, ${{\tilde \omega }_i}:{\mathbb{R}^{r_i}} \mapsto \mathbb{R}$, ${{\tilde \psi }_i}:{\mathbb{R}^{s_i}} \mapsto \mathbb{R}$, ${\omega _{i,j}}:{\mathbb{R}^{{r_{i,j}}}} \mapsto \mathbb{R}$ and ${\psi _{i,k}}:{\mathbb{R}^{{s_{i,k}}}} \mapsto \mathbb{R}$, respectively. 
${c_0},{c_1}, \cdots ,{c_m}$ are constant coefficients.
\end{definition}

\begin{definition}[Repeated variable, non-repeated variable, block and factor]
\label{def-4concepts}
In Eq. (\ref{eq-GSS}), the variables belong to ${X^r}$ and ${{\bar X}^r}$ are called repeated variables and non-repeated variables, respectively.
The sub-function ${\varphi _i}\left(  \cdot  \right)$ is called the $i$-th minimal block of $f\left( X \right)$, for $i = 1,2, \cdots ,m$.
Any combination of the minimal blocks is called a block of $f\left( X \right)$.
The sub-functions ${\omega _{i,j}}\left(  \cdot  \right)$ and ${\psi _{i,k}}\left(  \cdot  \right)$ are called the $j$-th and $k$-th factors of the repeated variables and non-repeated variables in $i$-th minimal block ${\varphi _i}\left(  \cdot  \right)$, respectively, for $j = 1,2, \cdots ,{p_i}$ and $k = 1,2, \cdots ,{q_i}$.
\end{definition}

\section{Model detection and determination}
\label{Section4}

In order to detect the separability of the GS system $f\left( X \right)$, we aim to divide $f\left( X \right)$ into a suitable number of minimal blocks, and further into factors as the typical Example \ref{ex2-3}. This technique can be considered as a generalized bi-correlation test (BiCT) method. The BiCT \cite{Luo2017} is developed to detect the separability of a certain additively or multiplicatively separable target function, i.e. the Eq. (\ref{def1eq}).

The modeling process of GS-system mainly includes two parts, namely inner optimization and outer optimization. The inner optimization will be invoked to determine the function model and coefficients of the factors ${{\omega _{i,j}}}$ and ${{\psi _{i,k}}}$. Fortunately, many state-of-the-art optimization techniques, e.g., parse-matrix evolution \cite{Luo2012-PME}, low dimensional simplex evolution \cite{Luo2012-LDSE}, artificial bee colony programming \cite{Karaboga2012}, etc. can all be easily used to optimize the factors. Then, the optimized factors of each minimal block are multiplied together to produce minimal blocks.

The outer optimization aims at combining the minimal blocks together with the proper global parameters $c_i$. The whole process  for modeling a GS system can be briefly described as follows:

\begin{enumerate}
\item (Minimal block detection) Partition a GS system into a number of minimal blocks with all the repeated variables fixed;
\item (Factor detection) Divide each minimal block into factors;
\item (Factor determination) Determine the factors by employing an optimization engine;
\item (Global assembling) Combine the optimized factors into minimal blocks multiplicatively, further into an optimization model linearly with proper global parameters.
\end{enumerate}

\begin{figure}[h]
\centering
\includegraphics[width=0.98\linewidth]{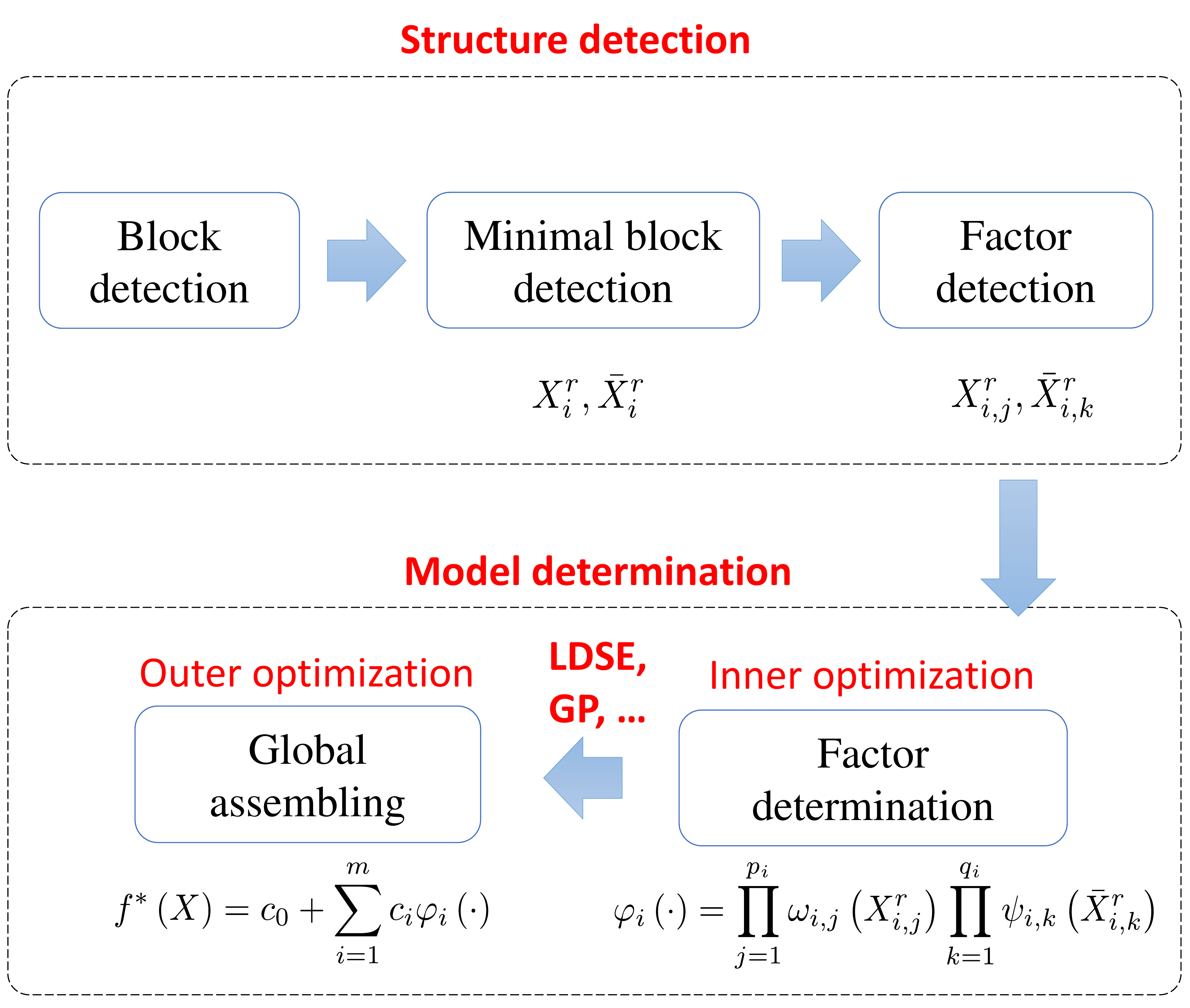}
\caption{Flowchart of modeling process.}
\label{workflow}
\end{figure}

The flowchart of the modeling process could be briefly illustrated in Fig. \ref{workflow}.

The proposed technique is described with functions with explicit expressions. While in practical applications, no explicit expression is available. In fact, for data-driven modeling problems, a surrogate model \cite{Forrester2009} of black-box type could be established as the underlying target function in advance.

\section{Numerical results and discussion}
\label{Section5}

In our implementation, a kind of global optimization method, low dimensional simplex evolution (LDSE) \cite{Luo2012-LDSE}, is chosen as the optimization engine. LDSE is a hybrid evolutionary algorithm for continuous global optimization. The performances including `structure optimization' and `coefficient optimization' capabilities of the proposed method are tested by comparing with a state-of-the-art software, Eureqa \cite{Schmidt2009}, which is a data-driven fitting tool based on genetic programming (GP). Eureqa was developed at the Computational Synthesis Lab at Cornell University by H. Lipson. 10 test cases are taken into account.

The calculation conditions are set as follows. The number of sampling points for each independent variable is 200. The regions for cases 1-5 and 7-10 are chosen as $[-3,3]$, while case 6 is $[1,3]$. The control parameters in LDSE are set as follows. The upper and lower bounds of fitting parameters is set as $-50$ and $50$. The population size $N_p$ is set to $N_p=10+10d$, where $d$ is the dimension of the problem. Sequence search and optimization method is suitable for global optimization strategy. The search will exit immediately if the mean square error is small enough (MSE $\leqslant {\varepsilon _{{\text{target}}}}$), and the tolerance (fitting error) is ${\varepsilon _{{\text{target}}}} = {10^{ - 6}}$. In order to reduce the effect of randomness, each test case is executed 20 times.

\begin{table*}
\scriptsize
\centering
\caption{10 test cases.}
\label{test_table_2}
\begin{tabular}{ll}
\hline\hline
No. & Target model                                                                     \\ \hline
1   & $f\left( {\mathbf{x}} \right) = 0.5 * {e^{{x_1}} * \sin 2{x_2}}$                                                              \\
2   & $f\left( {\mathbf{x}} \right) = 2 * \cos {x_1} + \sin \left( {3{x_2} - {x_3}} \right)$                                                        \\
3   & $f\left( {\mathbf{x}} \right) = 1.2 + 10 * \sin 2{x_1} - 3 * x_2^2 * \cos {x_3}$                                                                     \\
4   & $f\left( {\mathbf{x}} \right) = {x_3} * \sin {x_1} - 2 * {x_3} * \cos {x_2}$                                                                     \\
5   & $f\left( {\mathbf{x}} \right) = 2 * {x_1} * \sin {x_2} * \cos {x_4} - 0.5 * {x_4} * \cos {x_3}$                         \\
6   & $f\left( {\mathbf{x}} \right) = 10 + 0.2 * {x_1} - 0.2 * x_5^2 * \sin {x_2} + \cos {x_5} * \ln \left( {3{x_3} + 1.2} \right) - 1.2 * {e^{0.5{x_4}}}$                                 \\
7   & $f\left( {\mathbf{x}} \right) = 2 * {x_4} * {x_5} * \sin {x_1} - {x_5} * {x_2} + 0.5 * {e^{{x_3}}} * \cos {x_4}$                                                                \\
8   & $f\left( {\mathbf{x}} \right) = 1.2{\text{  +  }}2 * {x_4} * \cos {x_2} + 0.5 * {e^{1.2{x_3}}} * \sin 3{x_1} * \cos {x_4} - 2 * \cos \left( {1.5{x_5} + 5} \right)$                                                            \\
9   & $f\left( {\mathbf{x}} \right) = 0.5 * \frac{{\cos \left( {{x_3}{x_4}} \right)}}{{{e^{{x_1}}} * {x_2}^{2}}} * \sin \left( {1.5{x_5} - 2{x_6}} \right)$       \\
10  & $f\left( {\mathbf{x}} \right) = 1.2 - 2 * \frac{{{x_1} + {x_2}}}{{{x_3}}} * \cos {x_7} + 0.5 * {e^{{x_7}}} * {x_4} * \sin \left( {{x_5}{x_6}} \right)$                                                                    \\
 \hline\hline
\end{tabular}
\end{table*}

\begin{table*}
\scriptsize
\centering
\caption{Comparative results of the performances between the proposed method and Eureqa for modeling 10 test cases.}
\label{table2}
\begin{tabular}{llllllll|lll}
\hline\hline
\multirow{2}{*}{\begin{tabular}[c]{@{}l@{}}Case\\ No.\end{tabular}} & \multirow{2}{*}{Dim} & \multirow{2}{*}{\begin{tabular}[c]{@{}l@{}}No.\\ samples\end{tabular}} & \multicolumn{5}{l|}{Our method}                                                                                                                                                                                                                                                     & \multicolumn{3}{l}{Eureqa}                                                                                                     \\ \cline{4-11} 
                                                                    &                      &                                                                        & \begin{tabular}[c]{@{}l@{}}Repeated\\ variable\end{tabular} & \begin{tabular}[c]{@{}l@{}}No.\\ block\end{tabular} & \begin{tabular}[c]{@{}l@{}}No.\\ factor\end{tabular} & \begin{tabular}[c]{@{}l@{}}CPU\\ time\end{tabular} & MSE                                          & \begin{tabular}[c]{@{}l@{}}CPU\\ time\end{tabular} & MSE                                             & Remarks                 \\ \hline
1                                                                   & 2                    & 400                                                                    & None                                                        & 1                                                   & 2                                                    & \textbf{7s}                                      & $\leqslant {\varepsilon _{{\text{target}}}}$ & 7s                                                 & $\leqslant {\varepsilon _{{\text{target}}}}$    & Solutions are all exact \\
2                                                                   & 3                    & 600                                                                    & None                                                        & 2                                                   & 2                                                    & \textbf{9s}                                      & $\leqslant {\varepsilon _{{\text{target}}}}$ & $>$ 4m 12s                                             & $\left[ {0,2.33} \right] \times {10^{ - 8}}$ & 10 runs failed           \\
3                                                                   & 3                    & 600                                                                    & None                                                        & 2                                                   & 3                                                    & \textbf{9s}                                      & $\leqslant {\varepsilon _{{\text{target}}}}$ & $>$ 1m 9s                                              & $\leqslant {\varepsilon _{{\text{target}}}}$    & 2 runs failed           \\
4                                                                   & 3                    & 600                                                                    & $x_3$                                                       & 2                                                   & 4                                                    & \textbf{11s}                                     & $\leqslant {\varepsilon _{{\text{target}}}}$ & 55s                                                & $\leqslant {\varepsilon _{{\text{target}}}}$    & Solutions are all exact \\
5                                                                   & 4                    & 800                                                                    & $x_4$                                                       & 2                                                   & 5                                                    & \textbf{14s}                                     & $\leqslant {\varepsilon _{{\text{target}}}}$ & $>$ 2m 28s                                             & $\leqslant {\varepsilon _{{\text{target}}}}$    & 3 runs failed           \\
6                                                                   & 5                    & 1000                                                                   & $x_5$                                                       & 4                                                   & 6                                                    & \textbf{21s}                                     & $\leqslant {\varepsilon _{{\text{target}}}}$ & $\gg$ 6m 25s                                       & $\left[ {4.79,14.2} \right] \times {10^{ - 6}}$ & All runs failed         \\
7                                                                   & 5                    & 1000                                                                   & $x_4,x_5$                                                   & 3                                                   & 7                                                    & \textbf{16s}                                     & $\leqslant {\varepsilon _{{\text{target}}}}$ & $\gg$ 8m 38s                                       & $\left[ {4.05,7.68} \right] \times {10^{ - 4}}$ & All runs failed         \\
8                                                                   & 5                    & 1000                                                                   & $x_4$                                                       & 3                                                   & 6                                                    & \textbf{15s}                                     & $\leqslant {\varepsilon _{{\text{target}}}}$ & $\gg$ 6m 44s                                       & $\left[ {2.89,122.86} \right] \times {10^{ - 2}}$ & All runs failed         \\
9                                                                   & 6                    & 1200                                                                   & None                                                        & 1                                                   & 4                                                    & \textbf{9s}                                      & $\leqslant {\varepsilon _{{\text{target}}}}$ & $\gg$ 6m 59s                                       & $\left[ {1.4,8.54} \right] \times {10^{ - 1}}$ & All runs failed         \\
10                                                                  & 7                    & 1400                                                                   & $x_7$                                                       & 2                                                   & 6                                                    & \textbf{11s}                                     & $\leqslant {\varepsilon _{{\text{target}}}}$ & $\gg$ 6m 51s                                       & $\left[ {7.58,399.5} \right] \times {10^{ - 4}}$ & All runs failed         \\ \hline\hline
\end{tabular}
\end{table*}

The computing time (CPU time) consists three parts, $t=t_1+t_2+t_3$, where $t_1$ is for the separability detection, $t_2$ for factors modeling, and $t_3$ for global assembling. In \cite{Luo2017}, authors have demonstrated that both the separability detection and function recover processes are double-precision operations and thus cost much less time than the factor determination process. That is, $t \approx t_2$. It is very easy to see that the computational efficiency of the proposed method is higher than Eureqa's. Note that our method is executed on a single processor, while Eureqa is executed in parallel on 8 processors.

\section{Conclusion}
\label{Section6}
We have analyzed the different types of separability of some practical engineering problems and have established the mathematical model of the generalized separable system (GS system). In other to get the structure of the GS system, two types of variables in GS system have been identified, namely repeated variable and non-repeated variable. A new method, block and factor detection, has also been proposed to decompose the GS system into a number of block, further into minimal blocks and factors. The minimal blocks and factors are optimized and assembled with a global optimization search engine, low dimensional simplex evolution (LDSE). The proposed method is tested on several man-made test cases. Remarkable performance is concluded after comparing with a state-of-the-art data-driven fitting tool, Eureqa. Numerical results show the algorithm is effective, and can get the target function more rapidly and reliably.

\section*{Acknowledgment}
This work was supported by the National Natural Science Foundation of China (Grant No. 11532014).

\end{document}